\documentclass[times,twocolumn,final]{elsarticle}
\usepackage[caption=false,font=normalsize,labelfont=sf,textfont=sf]{subfig}
\usepackage{medima}
\usepackage{framed,multirow}
\usepackage{graphicx}
\usepackage{amstext}
\usepackage{amssymb}
\usepackage{latexsym}
\usepackage{threeparttable}
\usepackage{booktabs}
\usepackage{url}
\usepackage{xcolor}
\usepackage{hyperref}
\usepackage{float}
\definecolor{newcolor}{rgb}{.8,.349,.1}

\journal{Medical Image Analysis}

\begin{document}

\verso{H. Luo, H. Liu and K. Li \textit{et~al.}}

\begin{frontmatter}

\title{Automatic quality assessment for 2D fetal sonographic standard plane based on multi-task learning}%

\author[1]{Hong \snm{Luo}}

\author[2]{Han \snm{Liu}}

\author[3]{Kejun \snm{Li}}
\author[1]{Bo \snm{Zhang}\corref{cor1}}
\cortext[cor1]{Corresponding author.}
\ead{hxcszhangbo@163.com}

\address[1]{Department of Ultrasound, Sichuan University West China Second Hospital, Chengdu, 610066, China}
\address[2]{Glasgow College, University of Electronic Science and Technology of China, Chengdu, 611731, China}
\address[3]{Wangwang Technology Company, Chengdu, 610041, China}

\received{1 May 2013}
\finalform{10 May 2013}
\accepted{13 May 2013}
\availableonline{15 May 2013}
\communicated{S. Sarkar}

\begin{abstract}
The quality control of fetal sonographic (FS) images is essential for the correct biometric measurements and fetal anomaly diagnosis. However, quality control requires professional sonographers to perform and is often labor-intensive. To solve this problem, we propose an automatic image quality assessment scheme based on multi-task learning to assist in FS image quality control. An essential criterion for FS image quality control is that all the essential anatomical structures in the section should appear full and remarkable with a clear boundary. Therefore, our scheme aims to identify those essential anatomical structures to judge whether an FS image is the standard image, which is achieved by three convolutional neural networks. The Feature Extraction Network aims to extract deep level features of FS images. Based on the extracted features, the Class Prediction Network determines whether the structure meets the standard and Region Proposal Network identifies its position. The scheme has been applied to three types of fetal sections, which are the head, abdominal, and heart. The experimental results show that our method can make a quality assessment of an FS image within less a second. Also, our method achieves competitive performance in both the detection and classification compared with state-of-the-art methods.
\end{abstract}


\end{frontmatter}

\section{Introduction}
\begin{figure*}[h] 
	\begin{center} 
		\includegraphics[width=0.9\textwidth]{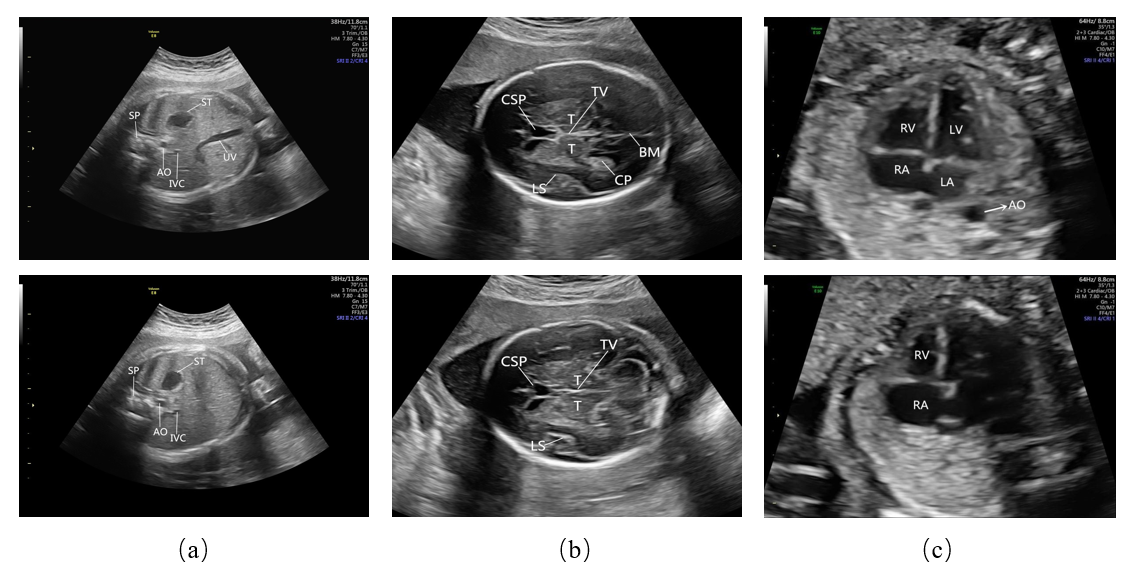} 
		\caption{Comparison of the standard plane (upper row) and non-standard plane (lower row) in three sections. (a)The lower abdominal FS image does not show the umbilical vein. (b) The lower head FS image does not show the brain midline and the choroid plexus. (c) The lower heart FS image does not show the left ventricle, left atrium, and descending aorta.
		} 
	\end{center}   
	\label{comp} 
\end{figure*}

\begin{figure*}[h] 
		\centering
		\includegraphics[width=0.9\textwidth]{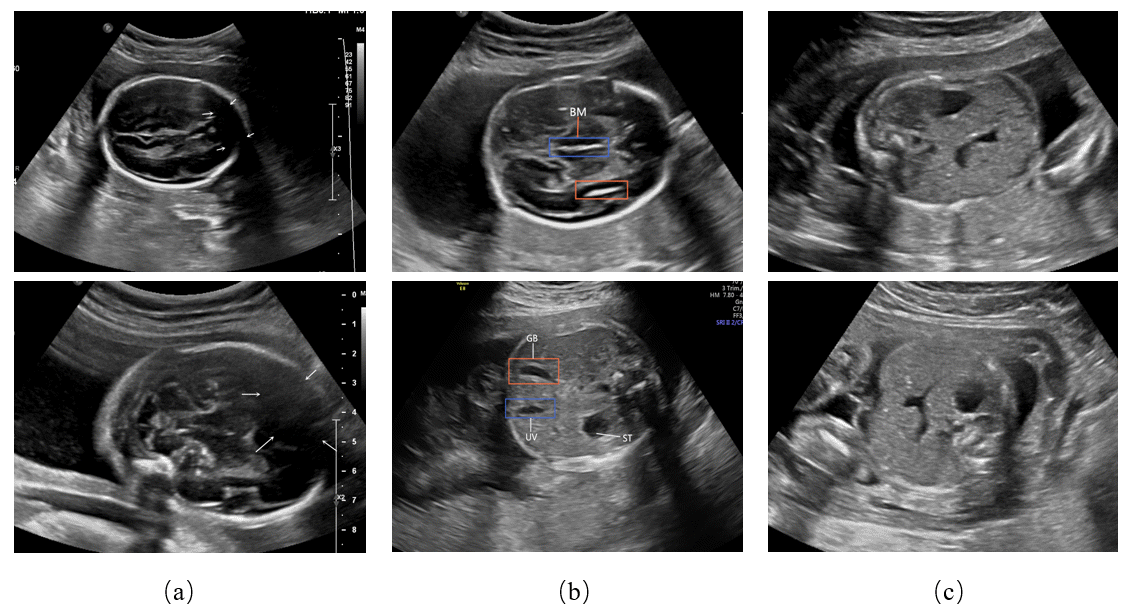} 
		\caption{  Illustration of different types of challenges. (a) The white arrows show the substantial interference of noise and shadowing effect. (b) In the upper graph, the blue box represents the real brain midline, and the orange box is the confusing anatomical structure with a similar shape. In the lower figure, the orange box represents the actual gallbladder, and the blue box represents the real umbilical vein. These two structures have a very similar shape. (c) FS images with different fetal positions, which will cause significant variations for the appearance of the images. 
		} 
  
	\label{chall} 
\end{figure*}

Fetal sonographic (FS) examinations are widely applied in clinical settings due to its non-invasive nature, reduced cost, and real-time acquisition \citep{Rueda}. FS examinations are consisted of first, second and third trimester examination, and limited examination \citep{AIUM}, which covers a range of critical inspections such as evaluation of a suspected ectopic pregnancy \citep{Chambers1990, Hilla}, and confirmation of the presence of an intrauterine pregnancy \citep{Barnhart2011,Jeve2011a,Thilaganathan2011}. The screening and evaluation of fetal anatomy are critical during the second and third trimester examinations. The screening is usually assessed by ultrasound after approximately 18 weeks’ gestational (menstrual) age. According to a survey \citep{Murphy2018}, neonatal mortality in the United States in 2016 was 5.9 deaths per 1,000 live births, and birth defects are the leading cause of infant deaths, accounting for 20\% of all infant deaths. Besides, congenital disabilities occur in one in every 33 babies (about 3\% of all babies) born in the United States each year. In this case, the screening and evaluation of fetal anomaly will provide crucial information to families prior to the anticipated birth of their child on diagnosis, underlying etiology, and potential treatment options, which can greatly improve the survival rate of the fetus. However, the physiological evaluation of fetal anomaly requires well trained and experienced sonographers to obtain standard planes. Although a detailed quality control guideline was developed for the evaluation of standard plane \citep{Murphy2018}, the accuracy of the measurements is highly dependent on the operator’s training, skill, and experience. According to a study \citep{Murphy2018}, intraobserver and interobserver variability exist in routine practice, and inconsistent image quality can lead to variances in specific anatomic structures captured by different operators. Furthermore, in areas where medical conditions are lagging, there is a lack of well-trained doctors, which makes FS examinations impossible to perform. To this end, automatic approaches for FS image quality assessment are needed to ensure that the image is captured as required by guidelines and provide accurate and reproducible fetal biometric measurements \citep{Zhang}. \\
\indent To obtain standard planes and assess the quality of FS images, it is necessary that all the essential anatomical structures in the imaging should appear full and remarkable with clear boundary \citep{AIUM}. For each medical section, there are different essential structures. In our research, we consider three medical sections: the heart section, the head section, and the abdominal section. The essential structures corresponding to these sections are given in Table \ref{tab1}. The list of essential anatomical structures used to evaluate the image quality is defined by the guideline \citep{AIUM} and further refined by two senior radiologists with more than ten years of experience of FS examination at the West China Second Hospital Sichuan University, Chengdu, China. A comparison of standard and non-standard planes can be illustrated in Fig. \ref{comp}.\\
\indent There are various types of challenges concerning the automatic quality control of FS images. As illustrated in Fig. \ref{chall}, the main challenges can be divided into three types: the first type is that the image usually suffers from the influence of noise and shadowing effect, the second type is that similar anatomical structures could be confused due to the low resolution of the images and the third type is that the fetal location during the scanning is unstable which will cause the rotation of some anatomical structure. The first type of challenges can only be solved by using more advanced scanning machines, but we can tackle the rest two challenges by a more scientific approach. Specifically, we need to find an efficient feature extraction method, which remains robust to the distinction between image rotation and similar structures. In recent years, deep learning techniques have been widely applied in many medical imaging fields due to the technique's stability and efficiency, such as anatomical object detection and segmentation \citep{Ghesu2016,Zhanga,Ghesu} and brain abnormalities detection \citep{Kebir2019,Sujit}. A well-designed neural network can efficiently extract the features for classification and identification. In our approach, we firstly design a Feature Extraction Network (FEN) to extract deep level features from FS images, then we feed the extracted features to the region proposal network (RPN) and then to class prediction network (CPN) to identify the region of interest (ROI) and classification simultaneously. Besides, to further improve the performance of our framework, we introduce a relation module to fully utilize the relationship between the entire image and each detected structure. In conclusion, our contribution can be summarized as follows:
\begin{itemize}
 	\item An automatic fetal sonographic image quality control framework is proposed for the detection and classification of the two-dimensional fetal heart standard plane. Our model is highly robust against the interference of image rotation and similar structures, and the detection speed is quite fast to meet the clinical requirements fully.
	\item We have introduced many recent advanced object detection technologies into our framework, such as relation module, spatial pyramid pooling, etc. The results of detection and classification are quite promising compared with state-of-the-art methods.
	\item Our framework is generalized and can be well applied to other standard planes. We have shown the results when our framework is applied to the abdominal and head standard plane, which are quite competitive compared with other existing advanced methods.
\end{itemize}

\begin{table}[htbp]
	\centering
	\caption{Essential anatomical structures for different sections.}
	\begin{tabular}{cl}
			\toprule[1.5pt]
			\multicolumn{1}{l}{Section name} & Essential anatomical structure \\
			\midrule
			\multirow{2}[14]{*}{Head section} & Cavum septi pellucidi \\
			& Thalamus \\ & Third ventricle \\ & Brain midline \\ & Lateral sulcus \\& Choroid plexus\\
			
			\cmidrule{1-1}    \multirow{3}[2]{*}{Abdominal section} & Stomach bubble \\
			& Spine \\
			& Umbilical vein \\
			& Aorta \\
			& Stomach \\
			\cmidrule{1-1}    \multirow{4}[2]{*}{Heart section} & Left ventricle \\
			& Left atrium \\
			& Right ventricle \\
			& Right atrium \\
			& Descending aorta \\
			\bottomrule[1.5pt]
		\end{tabular}
		\label{tab1}
\end{table}

\section{Related work}
In recent years, with the rapid development of computer vision technology, many intelligent automatic diagnostic techniques for FS images have been proposed. For example, Zehui Lin et al. \citep{Lin} proposed a multi-task convolutional neural network (CNN) framework to address the problem of standard plane detection and quality assessment of fetal head ultrasound images. Under the framework, they introduced prior clinical and statistical knowledge to reduce the false detection rate further. The detection speed of this method is quite fast, and the result achieves promising performance compared with state-of-the-art methods. Zhoubing Xu et al. \citep{Xu2018} proposed an integrated learning framework based on deep learning to perform view classification and landmark detection of the structures in the fetal abdominal ultrasound image simultaneously. The automatic framework achieved a higher classification accuracy better than clinical experts, and it also reduced landmark-based measurement errors.
Lingyun et al. \citep{Wu2017} proposed a computerized FS image quality assessment scheme to assist the quality control in the clinical obstetric examination of the fetal abdominal region. This method utilizes the local phase features along with the original fetal abdominal ultrasound images as input to the neural network. The proposed scheme achieved competitive performance in both view classification and region localization. Cheung-Wen Chang et al. \citep{Chang2018} proposed an automatic Mid-Sagittal Plane (MSP) assessment method for categorizing the 3D fetal ultrasound images. This scheme also analyzes corresponding relationships between resulting MSP assessments and several factors, including image qualities and fetus conditions. It achieves a correct high rate for the results of MSP detection. Chandan et al. \citep{Kumar} proposed an automatic method for fetal abdomen scan-plane identification based on three critical anatomical landmarks: the spine, stomach, and vein. In their approach, a Biometry Suitability Index (BSI) is proposed to judge whether the scan-plane can be used for biometry based on detected anatomical landmarks. The results of the proposed method 
over video sequences were closely similar to the clinical expert’s
assessment of scan-plane quality for biometry. Chen et al. \citep{Chen2015} presented transfer learning frameworks to the automatic detection of different standard planes from ultrasound imaging videos. The framework utilizes spatio-temporal feature learning with knowledge transferred recurrent neural network (T-RNN) consisting of a deep hierarchical visual feature extractor and a temporal sequence learning model. The experiment shows that its results outperform state-of-the-art methods. Baumgartner et al. \citep{Baumgartner} proposed a novel framework based on convolutional neural networks to automatically detect 13 standard fetal views in freehand 2-D ultrasound data and provide localization of the anatomical structures through a bounding box. A notable innovation is that the network learns to
localize the target anatomy using weak supervision based
on image-level labels only. Namburetea et al. \citep{Namburete2018} proposed a multi-task, fully convolutional neural network framework to address the problem of 3D fetal brain localization, alignment to a referential coordinate system, and structural segmentation. This method optimizes the network by learning features shared within the input data belonging to the correlated tasks, and it achieves a high brain overlap rate and low eye localization error. However, there are no existing automatic quality control methods for fetal heart planes, and the detection accuracy of existing methods on other planes is relatively low due to the use of the outdated design of neural networks. Therefore, it is desirable to propose a more efficient framework that can not only provide accurate clinical assessment in fetal heart plane but can also increase the detection accuracy in other planes.

\section{Methods}
\begin{figure*}[h] 
	\begin{center} 
		\includegraphics[width=0.9\textwidth]{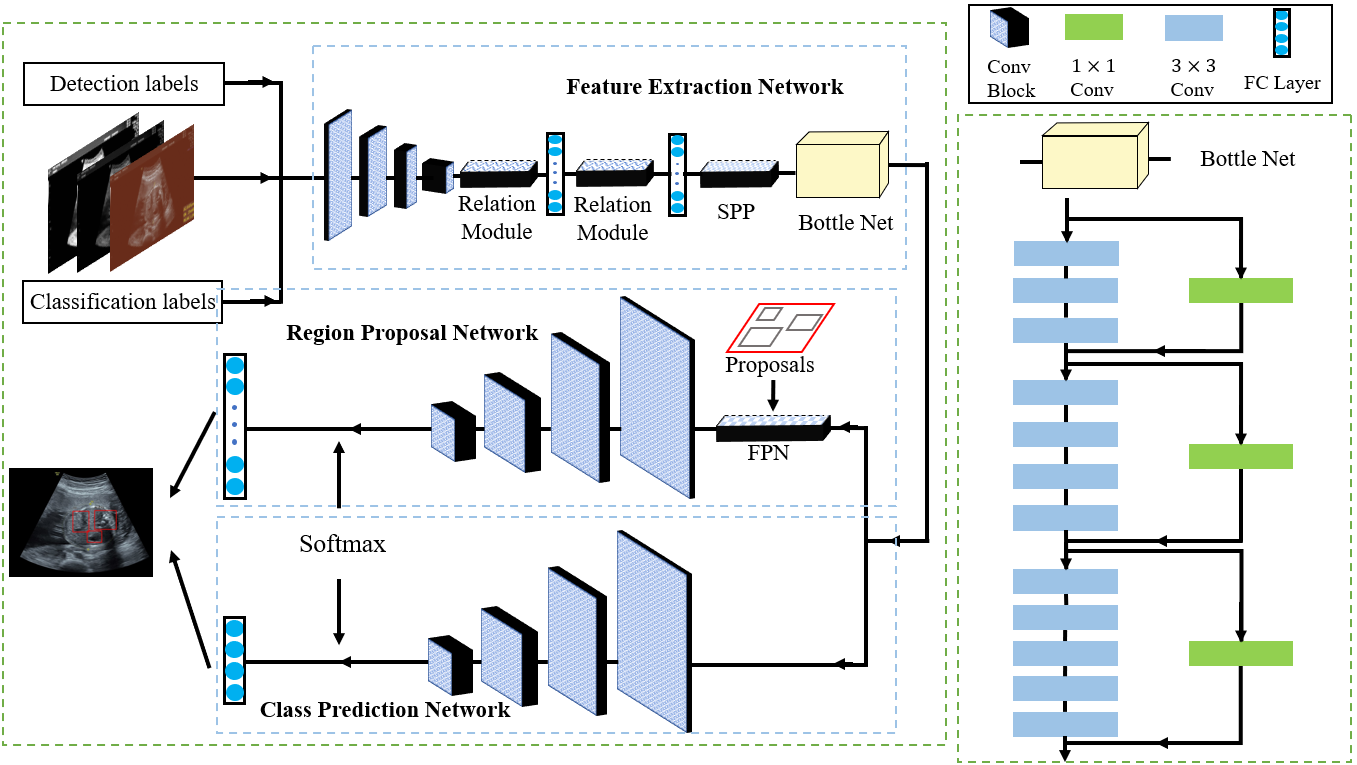} 
		\caption{The framework of our method. We train the network end-to-end to ensure the best performance. The framework contains three sections: Feature Extraction Network (FEN), Region Proposal Network (RPN), and Class Prediction Network (CPN). The FEN will help to extract the deep-level features of the image with the help of the relation module and Spatial Pyramid Pooling (SPP) layer, which is the input to RPN and CPN. The RPN will locate the position of essential structures based on the anchors generated by Feature Pyramid Network (FPN), and the CPN will help to judge and classify the structures. The final output will be a quality assessment of each essential structure and its location.
		} 
	\end{center}   
	\label{fig1} 
\end{figure*}

The framework of our methods can be illustrated in Fig. \ref{fig1}. First, the original image is smoothed by the Gaussian filter and inputted to FEN. Next, FEN will extract deep level features of the image by the convolutional neural network and input to RPN and CPN, respectively. Then, RPN will locate the position of essential structures with the help of feature pyramid networks, and CPN will judge whether the structures meet the standard as well as predict the class. Last, the two networks will combine information and output the final result. In this section, we will briefly introduce the network structure and then elaborate on the feature extraction, the ROI localization, and the structure classification in detail.

\subsection{Image Preprocessing} 
In this part, we mainly implement two operations. To protect the personal information of the subject in FS imaging, we firstly use the threshold method to remove the text in the image. Then we use Gaussian filtering to reduce image noise.\\
Gaussian filtering is a linear smoothing filter that is suitable for eliminating Gaussian noise and is widely used in the noise reduction process of image processing \citep{Deng}. Specifically, suppose $I$ denotes the original image matrix, then the processed image matrix $I_{\sigma}$ can be computed by:
\begin{equation} \label{eq1}
I_{\sigma}=I\ast G_{\sigma}
\end{equation}
\noindent where $G_{\sigma}$ is two-dimensional Gaussian kernel, and defined by:
\begin{equation} \label{eq2}
G_{\sigma}=1/(2\pi \sigma)e^{-\frac{x^2+y^2}{2\sigma^2}}
\end{equation}
The two-dimensional Gaussian function is rotationally symmetrical; that is, the smoothness of the filter in all directions is the same. This ensures that the Gaussian smoothing filter does not bias in either direction, which benefits the feature extraction in the following stage.

\subsection{Feature Extraction Network}
We use state of the art convolutional neural network techniques to design our feature extraction network. The convolutional neural network (CNN) has unique advantages in speech recognition and image processing with its special structure of local weight sharing, which can greatly reduce the number of parameters and improve the accuracy of recognition \citep{Ren,Zhao,Dai}. CNN typically consists of pairs of convolutional layers, average pooling layers, and fully connected (FC) layers. In the convolutional layer, several output feature maps can be obtained by the convolutional calculation between the input layer and kernel. Specifically, suppose $f_m^{n}$ denotes the $m$th output feature map in layer $n$, $f_k^{n-1}$ denotes the $k$th feature map in $n-1$ layer, $W_m^{n}$ denotes the kernel generating that feature map, then we can get:
\begin{equation} \label{eq3}
f_m^{n}=ReLu(\sum_{k=1}^{N}(W_m^{n}\ast f_k^{n-1})+b^n)
\end{equation}
where $b^n$ is the bias term in $n$th layer, $ReLu$ denotes  rectified linear unit, and is defined as: $ReLu(x)$ =$\max(x,0)$.
It is also worth mentioning that we use global average pooling (GAP) instead of local pooling for pooling layers. The aim is to apply GAP to replace the FC layer, which can regularize the structure of the entire network to prevent overfitting \citep{Lin2013}.
The setting of the convolution layer is shown in Table \ref{tab2}:
\begin{table}[htbp]
	\centering
	\caption{The setting of convolutional layer.}

		\begin{tabular}{lrrr}
			\toprule
			Layer & \multicolumn{1}{l}{Kernal size} & \multicolumn{1}{l}{Channel depth} & \multicolumn{1}{l}{Stride} \\
			\midrule
			C1    & 3     & 128   & 2 \\
			C2    & 3     & 256   & 2 \\
			C3    & 3     & 512   & 2 \\
			C4    & 3     & 1024  & 2 \\
			C5    & 3     & 2048  & 2 \\
			\bottomrule
		\end{tabular}%

	\label{tab2}%
\end{table}%

To fully utilize relevant features between objects and further improve detection accuracy, we introduce the relation module presented by Han Hu. Specifically, firstly the geometry weight is defined as:
\begin{equation} \label{eqa}
w_G^{mn}=max\{0,W_G\cdot\varepsilon_G(f_G^m,f_G^n)\}
\end{equation}
Where $f_G^m$ and $f_G^n$ are geometric features, $\varepsilon_G$ is  a dimension lifting transformation by using concatenation. After that, the appearance weight is defined as:
\begin{equation} \label{eqb}
w_A^{mn}=\frac{dot(W_Kf_A^m,W_Qf_A^n)}{\sqrt{d_k}}
\end{equation}
Where $W_K$ and $W_Q$ are the pixel weights from the previous network. Then the relation weight indicating the impact from other objects is computed as:
\begin{equation} \label{eqc}
w^{mn}=\frac{w_G^{mn}\cdot exp(w_A^{mn})}{\sum_kw_G^{kn}\cdot exp(w_A^{kn})}
\end{equation}
Lastly, the relation feature of the whole object set with respect to the $n^{th}$ object is defined as:
\begin{equation} \label{eqd}
f_R(n)=\sum_mw^{mn}\cdot(W_V\cdot f_A^m)
\end{equation}
This module achieves a great performance in the instance recognition and duplicate removal, which increases the detection accuracy significantly.\\
\indent The SPP layer we use here denotes the spatial pyramid pooling (SPP) layer presented by Kaiming He \citep{He}. Specifically, the response map after FC layer is divided into $1\times1$ (pyramid base), $2\times2 $ (lower middle of the pyramid), $4\times4$ (higher middle of the pyramid), $16\times16$ (pyramid top) four sub-maps and do max pooling separately. A problem with the traditional CNN network for feature extraction is that there is a strict limit on the size of the input image, this is because there is a need for the FC layer to complete the final classification and regression tasks, and since the number of neurons of the FC layer is fixed, the input image to the network must also have fixed size. Generally, there are two ways of fixing input image size: cropping and wrapping, but these two operations either cause the intercepted area not to cover the entire target or bring image distortion, thus applying SPP is necessary. The SPP network also contributes to multi-size extraction features and is highly tolerant to target deformation. \\
\indent The design of Bottle Net borrows the idea of Residual Networks \citep{He2016}. A common problem with deep networks is that gradient depth and gradient explosions are prone to occur as the depth deepens. The main reason for this phenomenon is the over-fitting problem caused by the loss of information. Each convolutional layer or pooling layer will downsample the image, producing a lossy compression effect. With network going deeper, some strange phenomena will appear in these images, where different categories of images produce similarly stimulating effects on the network. This reduction in the gap will make the final classification effect less than ideal. To let our network extract deeper features more efficiently, we add the residual network structure to our model. By introducing the data output of the previous layers directly into the input part of the latter data layer, we introduce a richer dimension by combining the original vector data and the subsequently downsampled data. In this way, the network can learn more features of the image.

\subsection{ROI Localization with RPN}
The RPN is designed to localize the ROI that encloses the essential structures given in Table \ref{tab1}. To achieve this goal, we first use a feature pyramid network (FPN) \citep{Lin2017a} to generate candidate anchors instead of the traditional RPN network used in Faster-RCNN \citep{Ren}. FPN could connect the high-level features of low-resolution and high-semantic information with the low-level features of high-resolution and low-semantic information from top to bottom so that features at all scales have rich semantic information. Specifically, the setting of FPN is shown in Table \ref{tab3}.\\
\begin{table}[htbp]
	\centering
	\caption{The setting of FPN.}

		\begin{tabular}{ccc}
			\toprule
			Pyramid level & Stride & Size \\
			\midrule
			3     & 8     & 32 \\
			4     & 16    & 64 \\
			5     & 32    & 128 \\
			6     & 64    & 256 \\
			7     & 128   & 512 \\
			\bottomrule
		\end{tabular}%

	\label{tab3}%
\end{table}%
\indent In the traning process, we define the metrics of intersection over union (IoU) to evaluate the quality of ROI localization:
\begin{equation} \label{eq4}
IoU = (A\cap B)/(A\cup B)
\end{equation}
where A is a computerized ROI and B is a manually labelled ROI (Ground Truth). In the training preocess, we set the samples with IoU higher than 0.5 as positive samples, and IoU lower than 0.5 as negative samples.

\subsection{Class Prediction with CPN }
For different sections, we use CPN to classify essential structures. For the head section, there are cavum septi pellucidi and thalamus to be classified. For the abdominal section, there are stomach bubble, spine, and umbilical vein to be classified. For the heart section, there are left ventricle, left atrium, right ventricle, and right atrium to be classified. To improve classification accuracy, we choose focal loss \citep{Lin2017} as the loss function. In the training process of the neural network, the internal parameters are adjusted by the minimization of the loss function of all training samples. The proposed focal loss enables highly accurate dense object detection in the presence of vast numbers of background examples, which is suitable in our model. The loss function can be defined as:
\begin{equation} \label{eq8}
FL(p_t)=-(1-p_t)^{\gamma}log(p_t)
\end{equation}
where $\gamma$ is the focusing parameter, and $\gamma \ge 0$. $p_t$ is defined as:

$$ p_t=\left\{
\begin{array}{lcl}
p       &      & {y=1}\\
1-p     &      & {\text{otherwise}}\\
\end{array} \right. $$

where $y$ represents the truth label of a sample, and $p$ represents the probability that the neural network predicts this class.

\section{Experiments and results}
In this section, we will start with a brief explanation of the process of obtaining and making data sets for training and testing. Then a systematic evaluation scheme will be proposed to test the efficacy of our method in FS examinations. The evaluation is carried out in four parts. First, we investigate the performance of ROI localization; we will use Mean Average Precision (mAP) and box-plot to evaluate it. Second, we quantitatively analyze the performance of classification with common indicators: accuracy (ACC), specificity (Spec), sensitivity (Sen), precision (Pre), F1-score (F1), and area of the receiver of operation curve (AUC). Third, we demonstrate the accuracy of our scheme when compared with experienced sonographers. Fourth, we test the running time of detecting a single FS image.

\subsection{Data preparation} 
All the FS images used for training and testing our model were acquired from the West China Second Hospital Sichuan University from April 2018 to January 2019. The FS images were recorded with a conventional hand-held 2-D FS probe on pregnant women in the supine position, by following the standard obstetric examination procedure. The fetal gestational ages of all subjects ranging from 20 to 34 weeks. All FS images were acquired with a GE Voluson E8 and Philips EPIQ 7 scanner.\\
\indent There are, in total, 1325 FS images of the head section, 1321 FS images of the abdominal section, and 1455 FS images of the heart section involved for the training and testing of our model. The training set, validation set, and test set of each section are all divided by a ratio of 3:1:1. The ROI labeling of essential structures in each section is achieved by two senior radiologists with more than ten years of experience in the FS examination by marking the smallest circumscribed rectangle of the positive sample. The negative ROI samples are randomly collected from the background of the images.

\subsection{Evaluation metrics}\
For testing the performance of ROI localization, firstly, we define the metrics of intersection over union (IoU) between prediction and ground truth and use box-plots to evaluate ROI localization intuitively. As illustrated before, IoU is defined as:
\begin{equation} \label{eq9}
IoU = (A\cap B)/(A\cup B)
\end{equation}
Where A is computerized ROI, and B is ground truth (manually labeled) ROI. Second, we use average precision (AP) to quantitively evaluate the detection results of each essential anatomical structure and mean average precision (mAP) to illustrate the overall quality of ROI localization.\\
To test the performance of classification results, we use several popular evaluation metrics. Suppose $TP$ represents the number of true positives of a certain class, $FP$ is the number of false positives, $FN$ is the number of false negatives and $TN$ is the number of true negatives, then the definitions of accuracy (ACC), specificity (Spec), sensitivity (Sen), precision (Pre) and F1-score (F1) are as following:
\begin{equation} \label{eq11}
ACC=\frac{TP+TN}{TP+FP+FN+TN}
\end{equation}

\begin{equation} \label{eq13}
Spec=\frac{TN}{FP+TN}
\end{equation}

\begin{equation} \label{eq12}
Sen=\frac{TP}{TP+FN}
\end{equation}

\begin{equation} \label{eq14}
Pre=\frac{TP}{TP+FP}
\end{equation}

\begin{equation} \label{eq15}
F1=\frac{2TP}{2TP+FN+FP}
\end{equation}
\indent The area of the receiver of operation curve (AUC) is defined as the area under the receiver operating characteristic (ROC) curve, which is equivalent to the probability that a randomly chosen positive example is ranked higher than a randomly chosen negative example \citep{Letters}.
To show the effectiveness of advanced techniques we add to the framework, and two different structures are also tested. Where NRM means the removal of the relation module, and NSPP means the removal of the Spatial Pyramid Pooling (SPP) layer in the feature extraction network. By comparing the difference in classification and detection results, it is clear to see their impact on overall network performance.
\subsection{Results of ROI localization}
\begin{figure*}[!t] \centering 
	\subfloat[Head section]{\includegraphics[width=2in]{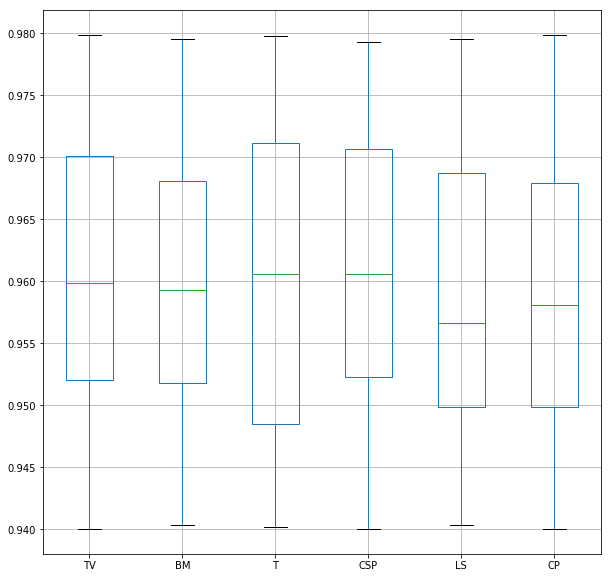} \label{fig_first_case}} 
	\hfil 
	\subfloat[Heart section]{\includegraphics[width=2in]{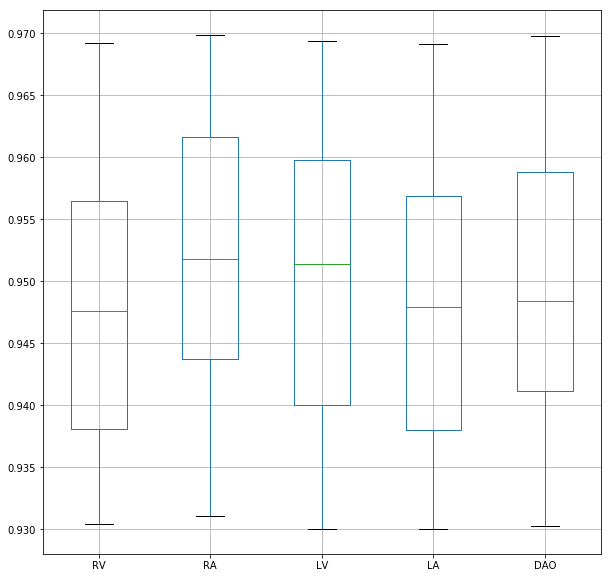} \label{fig_second_case}} 
	\hfil
	\subfloat[Abdominal section]{\includegraphics[width=2in]{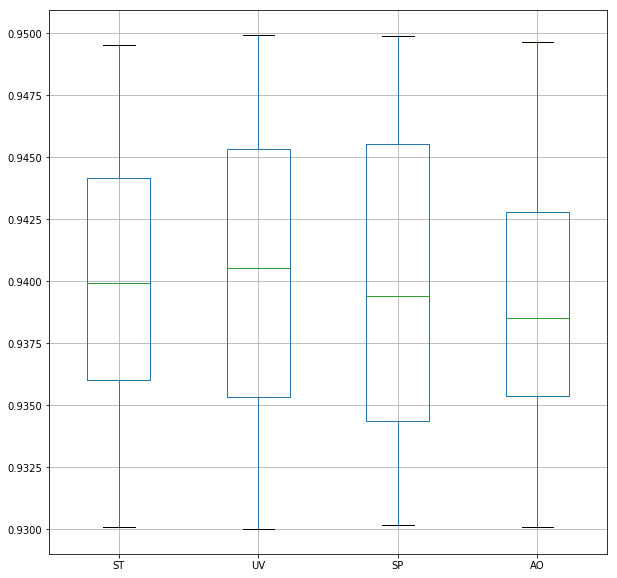} \label{fig_third_case}}
	\caption{Box-plots of IoU values for three sections. The three lines on each box represent the four quartiles of the IoU values.} 
	\label{fig3} 
\end{figure*}

To demonstrate the efficacy of our method in localizing the position of essential anatomical structures in FS images, we carry out the experimental evaluation in two parts. First, we use box-plots to evaluate ROI localization intuitively. Second, we use average precision (AP) and mean average precision (mAP) to illustrate the quality of ROI localization quantitively. \\
For the head standard plane, as illustrated in the related work, there is already a state-of-the-art method proposed for the quality assessment \citep{Lin} (Denoted as Lin), so we have compared its results with our method. Also, to show the effectiveness of advanced object detection techniques we add to the network, our methods have also been compared with other popular object detection frameworks, including SSD \citep{Liu2016}, YOLO \citep{Redmon2016,Redmon}, Faster R-CNN \citep{Ren}. The test of the effectiveness of the relation module we add to the network is also carried out, with Non-NM denoting the framework without the relation module.\\
\indent As shown in Fig. \ref{fig3}, our method has achieved a high IoU in all three sections. Specifically, for the head section, the median of IoU values in all the anatomical structures are above 0.955. Also, for the heart section and the abdominal section, the median is above 0.945 and 0.938, respectively. Also, the minimum of IoU values for all three sections are above 0.93. As a comparison, the state-of-the-art framework for the quality assessment of the fetal abdominal images proposed by Lingyun et al. \citep{Wu2017} has only achieved a median of below 0.9. It proves the effectiveness of our method in localizing ROI.\\
\indent As shown in Table \ref{head_comp}, we observe that our method has the highest mAP compared to the method proposed by Lin and other popular object detection frameworks. Also, we have improved the detection accuracy significantly in TV and CSP and overcome the limitation in Lin's method. This is because our method could detect flat and smaller anatomical structures more precisely. It is worth mentioning that after adding the relation module to our network, the detection accuracy has been significantly improved in all the anatomical structures, which proves the effectiveness of this module.\\
\begin{table*}[htb!]
	\centering
	\caption{Comparisons about detection results between our method and other methods in head section.}
	\begin{tabular}{llllllll}
		\toprule
		Method & TV    & BM    & T     & CP    & CSP   & LS    & mAP \\
		\midrule
		SSD   & 40.56 & 82.75 & 72.61 & 54.21 & 62.43 & 75.41 & 64.66 \\
		YOLOv2 & 35.43 & 79.31 & 38.56 & 62.7  & 83.67 & 83.31 & 63.83 \\
		Faster R-CNN VGG16 & 73.56 & 94.65 & 93.41 & 80.59 & 87.35 & 94.78 & 87.39 \\
		Faster R-CNN Resnet50 & 72.48 & 95.4  & 92.78 & 85.47 & 84.71 & 95.31 & 87.69 \\
		Lin   & 82.5  & 98.95 & 93.89 & 95.82 & 89.92 & 98.46 & 93.26 \\
		Non-NM & 71.42 & 94.38 & 89.92 & 82.45 & 86.78 & 92.45 & 86.23 \\
		Our Method & \textbf{86.12} & 98.87 & \textbf{94.21} & 93.76 & \textbf{95.57} & 97.92 & \textbf{94.41} \\
		\bottomrule
	\end{tabular}%
	\label{head_comp}%
\end{table*}%
\indent As shown in Table \ref{heart_comp}, since it is our first attempt to evaluate the image quality in the heart section, so we have only compared our method with state-of-the-art object detection frameworks. We observe that our approach has the highest average precision in all the anatomical structures. Also, as shown in Table \ref{ab_comp}, we have achieved quite promising detection accuracy. It proves that our framework is generalized and can be well applied to the quality assessment of other standard planes.

\begin{table*}[htb!]
	\centering
	\caption{Comparisons about detection results between our method and other methods in heart section.}
	\begin{tabular}{lllllll}
		\toprule
		Method & RV    & RA    & LV    & LA    & DAO   & mAP \\
		\midrule
		SSD   & 60.27 & 62.43 & 67.61 & 54.21 & 74.67 & 63.84 \\
		YOLOv2 & 71.31 & 69.39 & 71.56 & 62.7  & 90.74 & 73.14 \\
		Faster R-CNN VGG16 & 85.52 & 81.15 & 87.11 & 80.59 & 95.57 & 85.99 \\
		Faster R-CNN Resnet50 & 89.44 & 83.59 & 90.78 & 85.47 & 94.01 & 88.66 \\
		Non-NM & 84.35 & 85.43 & 91.23 & 82.45 & 95.86 & 87.86 \\
		Our Method & \textbf{93.03} & \textbf{95.71} & \textbf{95.65} & \textbf{93.76} & \textbf{97.34} & \textbf{95.10} \\
		\bottomrule
	\end{tabular}%
	\label{heart_comp}%
\end{table*}%

\begin{table*}[htb!]
	\centering
	\caption{Comparisons about detection results between our method and other methods in abdominal secton.}
	\begin{tabular}{llllll}
		\toprule
		Method & ST    & UV    & SP    & AO    & mAP \\
		\midrule
		SSD   & 80.74 & 83.43 & 76.23 & 62.13 & 75.63 \\
		YOLOv2 & 88.21 & 85.47 & 78.61 & 67.71 & 80 \\
		Faster R-CNN VGG16 & 90.25 & 92.15 & 88.72 & 82.54 & 88.42 \\
		Faster R-CNN Resnet50 & 91.29 & 93.59 & 90.85 & 81.34 & 89.27 \\
		Non-NM & 93.41 & 91.76 & 94.38 & 90.34 & 92.47 \\
		Our Method & \textbf{97.33} & \textbf{97.77} & \textbf{96.25} & \textbf{94.16} & \textbf{96.38} \\
		\bottomrule
	\end{tabular}%
	\label{ab_comp}%
\end{table*}%
\subsection{Results of classification accuracy}
\begin{figure}[!htb] 
		\centering

 		\includegraphics[width=0.48\textwidth]{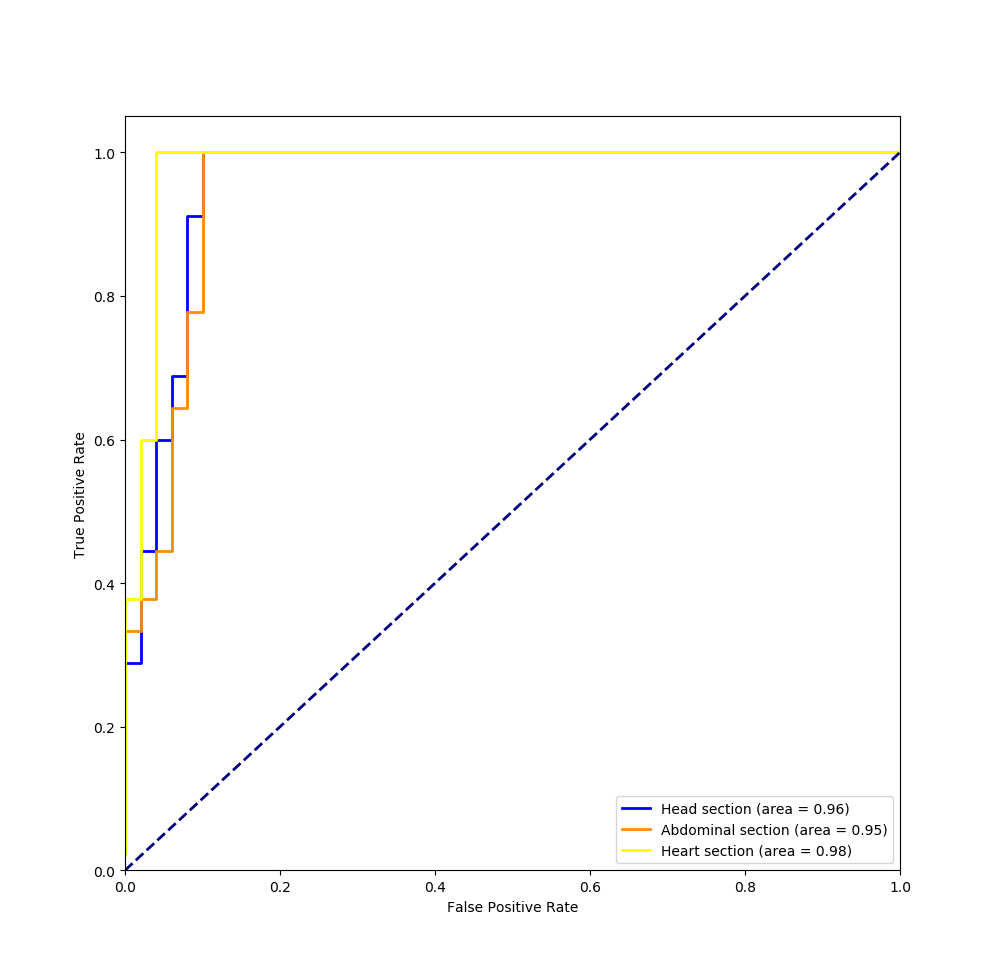} 
 		\centering
 		\caption{ROC curves of classification results in three sections
 		} 
  	 	\label{ROC_curve} 
 
 \end{figure}

To illustrate the performance of our model in classifying the essential anatomical structures, we firstly use area of receiver (ROC) of operation curve to characterize the performance of the classifier visually, then we use several authoritative indicators to measure it quantitatively: accuracy (ACC), specificity (Spec), sensitivity (Sen), precision (Pre) and F1-score (F1). Also, to show the effectiveness of our proposed network in classification, we have compared our method with other popular classification networks, including AlexNet \citep{Krizhevsky2012}, VGG16, VGG19 \citep{Simonyan2014}, and ResNet50 \citep{He2016a}. The comparison with Lin's method is also carried out.\\
\indent As shown in Fig. \ref{ROC_curve}, it is observed that the classifier achieves quite promising performance in all the three sections with the true positive rate reaching 100\% while the false positive rate is less than 10\%. Also, the ROC achieves at 0.96, 0.95, and 0.98 for the head section, abdominal section, and heart section, respectively. \\
From Table \ref{class_head}, we can observe that the classification results of our method are superior to other state-of-the-art methods. Specially, we achieve the best results with a precision of 94.63\%, a specificity of 96.39\%, and an AUC of 98.26\%, which are better than Lin's method. The relative inferior results in sensitivity, accuracy, and F1-score can be further improved if we add prior clinical knowledge into our framework \citep{Lin}. Table \ref{class_abdominal} and \ref{class_heart} illustrate the classification results in abdominal and heart section. We can observe that our method has achieved quite promising results in most indicators compared with existing methods. It demonstrates the effectiveness of our proposed method in classifying anatomical structures of all the sections.

\begin{table}[htbp!]
	\centering
	\caption{Comparisons about classification results between our method and other methods in head section.}
	 \setlength{\tabcolsep}{1.6mm}{
	\begin{tabular}{lllllll}
		\toprule
		Indicator & Prec  & Sen   & ACC   & F1    & Spec  & AUC \\
		\midrule
		AlexNet & 90.21 & 91.28 & 92.22 & 91.77 & 92.46 & 95.29 \\
		VGG16 & 92.34 & 92.48 & 91.79 & 90.68 & 93.71 & 96.37 \\
		VGG19 & 94.31 & 93.72 & 92.91 & 91.01 & 94.33 & 97.14 \\
		ResNet50 & 94.36 & 93.01 & 93.96 & \textbf{94.21} & 92.91 & 97.59 \\
		Lin   & 93.57 & \textbf{93.57} & \textbf{94.37} & 93.57 & 95.00    & 98.18 \\
		Our method & \textbf{94.63} & 92.41 & 94.31 & 93.17 & \textbf{96.39} & \textbf{98.26} \\
		\bottomrule
	\end{tabular}}
	\label{class_head}%
\end{table}%

\begin{table}[htbp!]
	\centering
	\caption{Comparisons about classification results between our method and other methods in abdominal section.}
	\setlength{\tabcolsep}{1.6mm}{
	\begin{tabular}{lllllll}
		\toprule
		Indicator & Prec  & Sen   & ACC   & F1    & Spec  & AUC \\
		\midrule
		AlexNet & 91.34 & 91.24 & 92.11 & 90.43 & 91.53 & 93.12 \\
		VGG16 & 93.21 & 93.59 & 91.33 & 90.91 & 92.23 & 92.76 \\
		VGG19 & 93.42 & \textbf{94.27} & 92.12 & \textbf{93.91} & 92.91 & 93.21 \\
		ResNet50 & 94.52 & 94.10  & 93.66 & 93.76 & 95.92 & 96.32 \\
		Our method & \textbf{95.67} & 93.56 & \textbf{96.31} & 92.34 & \textbf{97.92} & \textbf{98.54} \\
		\bottomrule
	\end{tabular}}
	\label{class_abdominal}%
\end{table}%

\begin{table}[htbp!]
	\centering
	\caption{Comparisons about classification results between our method and other methods in heart section.}
	\setlength{\tabcolsep}{1.6mm}{
	\begin{tabular}{lllllll}
		\toprule
		Indicator & Prec  & Sen   & ACC   & F1    & Spec  & AUC \\
		\midrule
		AlexNet & 90.34 & 91.44 & 92.32 & 90.78 & 91.28 & 93.87 \\
		VGG16 & 92.76 & 93.71 & 92.13 & 93.12 & 92.33 & 93.43 \\
		VGG19 & 93.68 & 94.31 & 93.37 & 92.88 & 93.54 & 94.41 \\
		ResNet50 & 95.91 & 94.31 & \textbf{93.52} & 93.45 & 94.42 & 93.56 \\
		Our method & \textbf{96.71} & \textbf{94.73} & 93.32 & \textbf{95.91} & \textbf{94.49} & \textbf{95.67} \\
		\bottomrule
	\end{tabular}}
	\label{class_heart}%
\end{table}%
\subsection{Running Time Analysis}
\begin{figure}[!htb]
	\centering
	\includegraphics[width=0.48\textwidth]{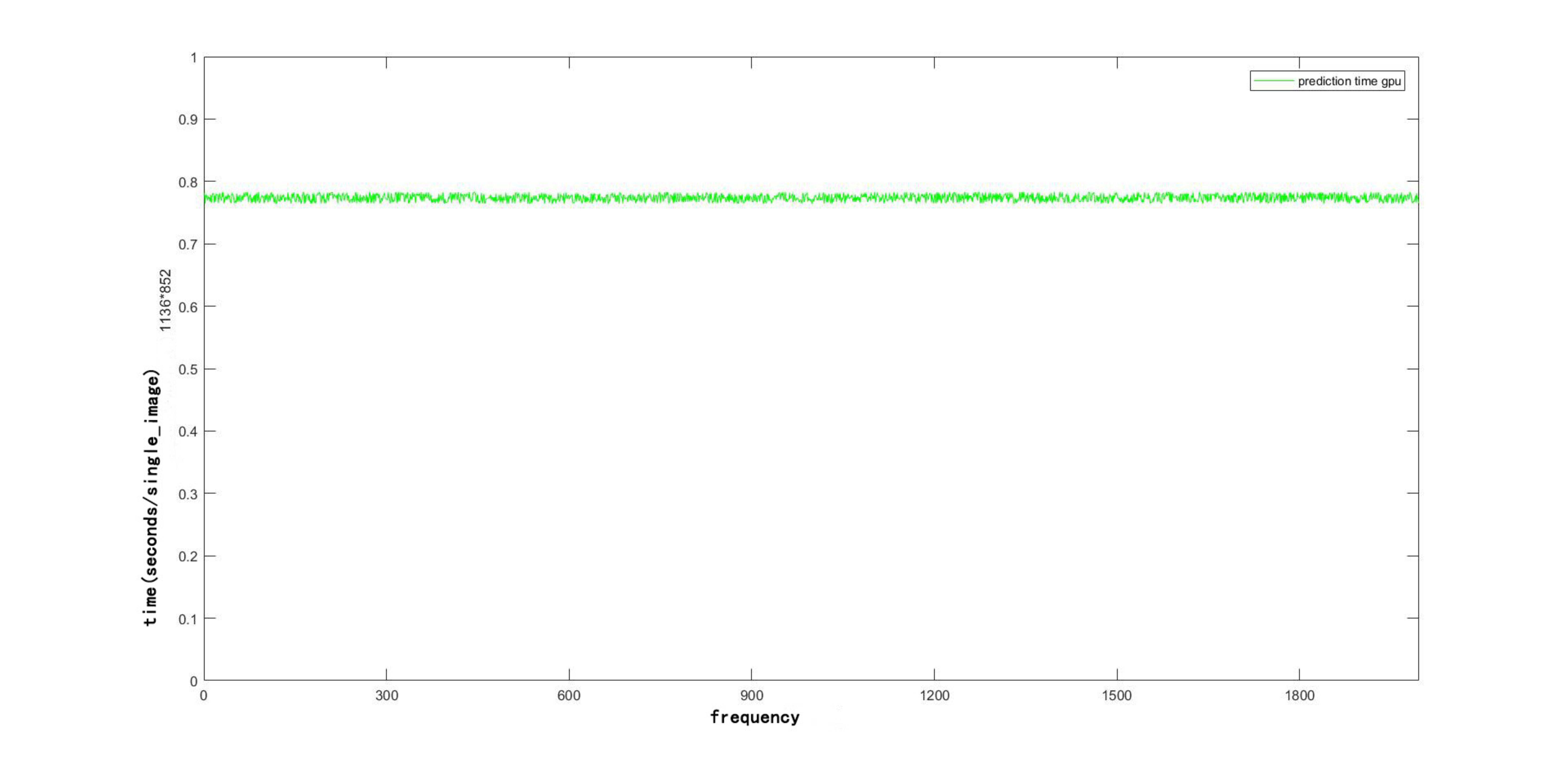}
	\caption{Running time of detecting single FS image. The abscissa represents the number of tests.}
	\label{fig4}
\end{figure}

First, We test the running time of detecting a single FS image in a workstation equipped with 3.60 GHz Intel Xeon E5-1620 CPU and a GP106-100 GPU. The results are given in Fig. \ref{fig4}. It is observed that detecting a single frame could only cost 0.78s, which is fast enough to meet clinical needs. As shown in Table \ref{time}, we have compared our method with different single-task and multi-task networks in terms of the average detection speed and network parameters. It is observed that although the network parameters of our method are much more than Faster R-CNN + ResNet50, there is not much difference in detection time, this is because our network shared many low-level features, which could achieve a more efficient detection with using only a few parameters.\\
\begin{table}[ht]
	\centering\small
	\caption{The detection speed and parameters of different single-task and multi-task methods}
	\setlength{\tabcolsep}{1.3mm}{
	\begin{tabular}{llll}
		\toprule
		& Method & Speed(s) & Parameters(M) \\
		\midrule
		Single-task & AlexNet & 0.015 & 26.01 \\
		& VGG16 & 0.02  & 67.04 \\
		& VGG19 & 0.02  & 72.1 \\
		& ResNet50 & 0.053 & 22.5 \\
		Multi-task & Faster R-CNN VGG16 & 0.534 & 130.45 \\
		& Faster R-CNN ResNet50 & 0.586 & 27.07 \\
		& Our Method & 0.78  & 62.38 \\
		\bottomrule
	\end{tabular}}
	\label{time}%
\end{table}%

\begin{figure*}[htbp!] 
	\centering
	\includegraphics[width=0.8\textwidth]{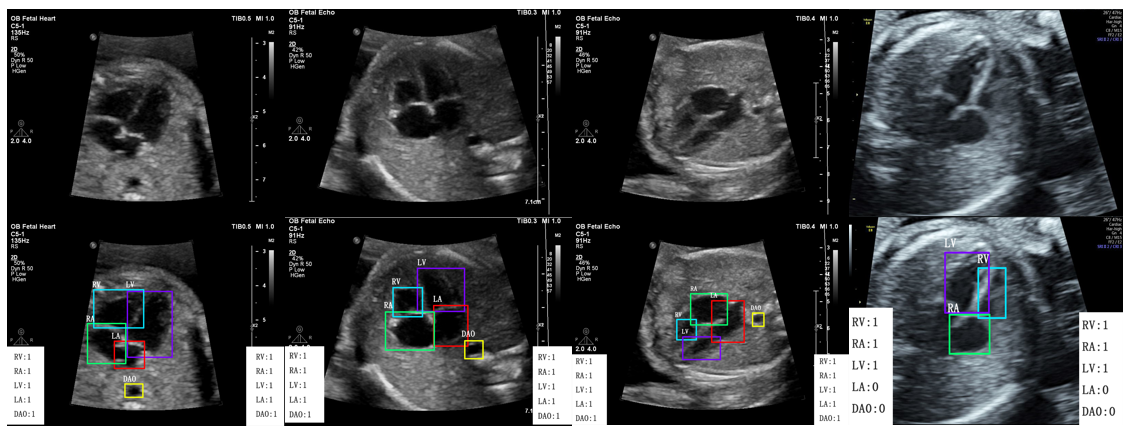} 
	\caption{Demonstration that our results perfectly match with the annotations of ground truth in the heart section. The classification results in the left white box are the ground truth labeled by professional radiologists, and the results in the right white box are the detection results of our method. '1' means the anatomical structure meets the quality requirement, and '0' means the structure does not meet the requirement.} 
	
	\label{fig5} 
\end{figure*}

\begin{figure*}[htbp!] 
	\centering
	\includegraphics[width=0.8\textwidth]{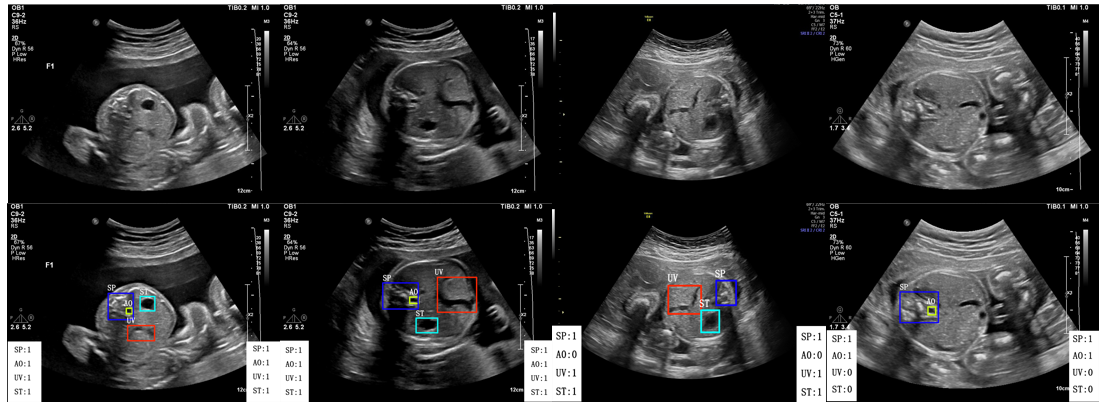} 
	\caption{Demonstration that our results perfectly match with the annotations of ground truth in the abdominal section. The classification results in the left white box are the ground truth labeled by professional radiologists, and the results in the right white box are the detection results of our method. '1' means the anatomical structure meets the quality requirement, and '0' means the structure does not meet the requirement.} 
	
	\label{fig6} 
\end{figure*}

\begin{figure*}[htbp!] 
	\centering
	\includegraphics[width=0.8\textwidth]{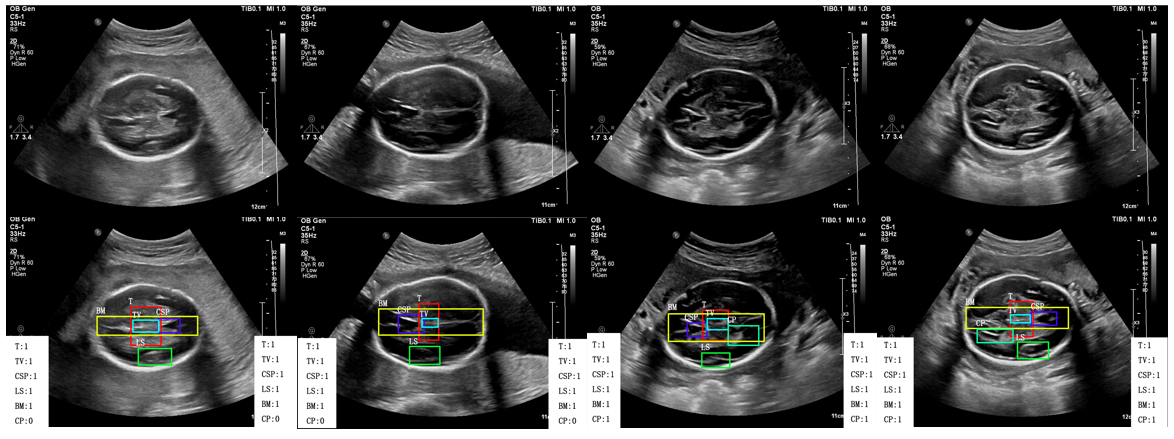} 
	\caption{Demonstration that our results perfectly match with the annotations of ground truth in the head section. The classification results in the left white box are the ground truth labeled by professional radiologists, and the results in the right white box are the detection results of our method. '1' means the anatomical structure meets the quality requirement, and '0' means the structure does not meet the requirement.} 
	
	\label{fig7} 
\end{figure*}

\indent Fig. \ref{fig5}, Fig. \ref{fig6} and Fig. \ref{fig7} depict the comparison of our results with the manually labeled images by experts in the head section, abdominal section, and heart section, respectively. Our method displays the classification and detection results simultaneously to assist in sonographers' observation. It can be seen that our method is perfectly aligned with professional sonographers. 

\section{Discussions}
In this paper, an autonomous image quality assessment approach for FS images was investigated. The experimental results show that our proposed scheme achieves a highly precise ROI localization and a considerable degree of accuracy for all the essential anatomical structures in the three standard planes. Also, the conformance test shows that our results are highly consistent with those of professional sonographers, and running time tests show that the image detection speed per frame is much higher than sonographers, which means this scheme can effectively replace the work of sonographers. In our proposed network, to further improve detection and classification accuracy, we also modify the recently published advanced object detection technologies and adapt them to our model. The experiment shows these modules are highly useful, and the overall performance is better than the state-of-the-art methods such as the FS image assessment framework proposed by Zehui Lin \citep{Lin}. After the Feature Extraction Network, we also divide the network into Region Proposal Network and Class Prediction Network. Accordingly, the features in the detection network can avoid interfering with the features in the classification network, so the detection accuracy is further increased. Also, the detection speed can be significantly improved, as the classification and localization are performed simultaneously. \\
Although our method achieves quite promising results, there are still some limitations. First, for the training sets, we regard the manually labeled FS images by two professional sonographers as the ground truth, but the results of manual labeling will have some accidental deviation even though they all have more than ten years of experience. In future studies, we will invite more professional clinical expects to label the FS images and collect more representative datasets. Second, there still remain some detection and classification errors in our results. This is because our evaluation criteria are rigorous, and the midsection of a single anatomical structure could lead to a negative score on the image. Third, all the FS images are collected from GE Voluson E8 and Philips EPIQ 7 scanner, however, different types of ultrasonic instruments will produce different ultrasound images, which may cause our method not to be applied well to the FS images produced by other machines.\\
Our proposed method further boosts the accuracy in the assessment of two dimensional FS standard plane. Although three dimensional and four-dimensional ultrasound testing are popular recently, they are mainly utilized to meet the needs of pregnant women and their families to view baby pictures instead of serving the diagnosing purpose visually. Two-dimensional ultrasound images are still the most authoritative basis for judging fetal development \citep{AIUM}. As illustrated before, there are still many challenges for the automatic assessment of 2D ultrasound images, such as shadowing effects, similar anatomical structures, different fetal positions, etc. To overcome these challenges and futher promote the accuracy and robustness of detection and classification, it may be useful to add some prior clinical knowledge \citep{Lin} and more advanced attention modules to the network. In the future, we will also investigate the automatic selection technology for finding the standard scanning plane, which will find a standard plane containing all the essential anatomical structures without sonographers' intervention.
\section{Conclusions}
The quality control for the FS images is significant for the biometric measurements and fetal anomaly diagnosis. However, the current FS examinations require well-trained and experienced sonographers to perform and are full of subjectivity. To develop an autonomous quality control approach and offer objective and accurate assessments, we propose an automatic FS image quality control scheme in this paper. In our scheme, we have designed three networks for detection and classification based on deep learning. The proposed scheme can reduce the workload of doctors while ensuring nearly the same accuracy and lower the skill requirement of sonographers, which will make it possible for fetal FS examinations in those areas where medical conditions are lagging. \\
Our proposed scheme has been evaluated with extensive experiments. The results show that our scheme is not only comparable to manual labeling by experts in locating the location of anatomical structures, but also very accurate on the classification. Furthermore, evaluating an FS image takes less than a second. In the future, we will extend our research into more fetal sections and try to propose an automatic selection technology for the standard plane of the FS images.

\section*{Declaration of Competing Interest}
The authors have no financial interests or personal relationships that may cause a conflict of interest.

\section*{Acknowledgments}
We acknowledge West China Second Hospital Sichuan University for providing the fetal ultrasound image datasets. This research did not receive any specific grant from funding agencies in the public, commercial, or
not-for-profit sectors.
\newpage

\bibliographystyle{model2-names.bst}
\biboptions{authoryear}
\bibliography{medima-template}

\begin{thebibliography}{38}
\expandafter\ifx\csname natexlab\endcsname\relax\def\natexlab#1{#1}\fi
\providecommand{\url}[1]{\texttt{#1}}
\providecommand{\href}[2]{#2}
\providecommand{\path}[1]{#1}
\providecommand{\DOIprefix}{doi:}
\providecommand{\ArXivprefix}{arXiv:}
\providecommand{\URLprefix}{URL: }
\providecommand{\Pubmedprefix}{pmid:}
\providecommand{\doi}[1]{\href{http://dx.doi.org/#1}{\path{#1}}}
\providecommand{\Pubmed}[1]{\href{pmid:#1}{\path{#1}}}
\providecommand{\bibinfo}[2]{#2}
\ifx\xfnm\relax \def\xfnm[#1]{\unskip,\space#1}\fi
\bibitem[{AIUM(2013)}]{AIUM}
\bibinfo{author}{AIUM}, \bibinfo{year}{2013}.
\newblock \bibinfo{title}{{AIUM practice guideline for the performance of
  obstetric ultrasound examinations}}.
\newblock \DOIprefix\doi{10.7863/ultra.32.6.1083}.
\bibitem[{Barnhart et~al.(2011)Barnhart, {Van Mello}, Bourne, Kirk, {Van
  Calster}, Bottomley, Chung, Condous, Goldstein, Hajenius, Mol, Molinaro,
  {O'Flynn O'Brien}, Husicka, Sammel and Timmerman}]{Barnhart2011}
\bibinfo{author}{Barnhart, K.}, \bibinfo{author}{{Van Mello}, N.M.},
  \bibinfo{author}{Bourne, T.}, \bibinfo{author}{Kirk, E.},
  \bibinfo{author}{{Van Calster}, B.}, \bibinfo{author}{Bottomley, C.},
  \bibinfo{author}{Chung, K.}, \bibinfo{author}{Condous, G.},
  \bibinfo{author}{Goldstein, S.}, \bibinfo{author}{Hajenius, P.J.},
  \bibinfo{author}{Mol, B.W.}, \bibinfo{author}{Molinaro, T.},
  \bibinfo{author}{{O'Flynn O'Brien}, K.L.}, \bibinfo{author}{Husicka, R.},
  \bibinfo{author}{Sammel, M.}, \bibinfo{author}{Timmerman, D.},
  \bibinfo{year}{2011}.
\newblock \bibinfo{title}{{Pregnancy of unknown location: A consensus statement
  of nomenclature, definitions, and outcome}}.
\newblock \bibinfo{journal}{Fertility and Sterility} \bibinfo{volume}{95},
  \bibinfo{pages}{857--866}.
\newblock \DOIprefix\doi{10.1016/j.fertnstert.2010.09.006}.
\bibitem[{Baumgartner et~al.(2017)Baumgartner, Kamnitsas, Matthew, Fletcher,
  Smith, Koch, Kainz and Rueckert}]{Baumgartner}
\bibinfo{author}{Baumgartner, C.F.}, \bibinfo{author}{Kamnitsas, K.},
  \bibinfo{author}{Matthew, J.}, \bibinfo{author}{Fletcher, T.P.},
  \bibinfo{author}{Smith, S.}, \bibinfo{author}{Koch, L.M.},
  \bibinfo{author}{Kainz, B.}, \bibinfo{author}{Rueckert, D.},
  \bibinfo{year}{2017}.
\newblock \bibinfo{title}{{SonoNet: Real-Time Detection and Localisation of
  Fetal Standard Scan Planes in Freehand Ultrasound}}.
\newblock \bibinfo{journal}{IEEE Transactions on Medical Imaging}
  \bibinfo{volume}{36}, \bibinfo{pages}{2204--2215}.
\newblock \URLprefix
  \url{https://ieeexplore.ieee.org/abstract/document/7974824/},
  \DOIprefix\doi{10.1109/TMI.2017.2712367},
  \href{http://arxiv.org/abs/1612.05601}{\tt arXiv:1612.05601}.
\bibitem[{Chambers et~al.(1990)Chambers, Muir and Haddad}]{Chambers1990}
\bibinfo{author}{Chambers, S.E.}, \bibinfo{author}{Muir, B.B.},
  \bibinfo{author}{Haddad, N.G.}, \bibinfo{year}{1990}.
\newblock \bibinfo{title}{{Ultrasound evaluation of ectopic pregnancy including
  correlation with human chorionic gonadotrophin levels}}.
\newblock \bibinfo{journal}{British Journal of Radiology} \bibinfo{volume}{63},
  \bibinfo{pages}{246--250}.
\newblock \DOIprefix\doi{10.1259/0007-1285-63-748-246}.
\bibitem[{Chang et~al.(2018)Chang, Huang, Huang, Sun and Tsai}]{Chang2018}
\bibinfo{author}{Chang, C.W.}, \bibinfo{author}{Huang, S.T.},
  \bibinfo{author}{Huang, Y.H.}, \bibinfo{author}{Sun, Y.N.},
  \bibinfo{author}{Tsai, P.Y.}, \bibinfo{year}{2018}.
\newblock \bibinfo{title}{{Categorizating 3d fetal ultrasound image database in
  first trimester pregnancy based on mid-sagittal plane assessments}}, in:
  \bibinfo{booktitle}{Proceedings - Applied Imagery Pattern Recognition
  Workshop}, \bibinfo{publisher}{Institute of Electrical and Electronics
  Engineers Inc.}
\newblock \DOIprefix\doi{10.1109/AIPR.2017.8457976}.
\bibitem[{Chen et~al.(2015)Chen, Dou, Ni, Cheng, Qin, Li and Heng}]{Chen2015}
\bibinfo{author}{Chen, H.}, \bibinfo{author}{Dou, Q.}, \bibinfo{author}{Ni,
  D.}, \bibinfo{author}{Cheng, J.Z.}, \bibinfo{author}{Qin, J.},
  \bibinfo{author}{Li, S.}, \bibinfo{author}{Heng, P.A.}, \bibinfo{year}{2015}.
\newblock \bibinfo{title}{{Automatic fetal ultrasound standard plane detection
  using knowledge transferred recurrent neural networks}}, in:
  \bibinfo{booktitle}{Lecture Notes in Computer Science (including subseries
  Lecture Notes in Artificial Intelligence and Lecture Notes in
  Bioinformatics)}. \bibinfo{publisher}{Springer Verlag}. volume
  \bibinfo{volume}{9349}, pp. \bibinfo{pages}{507--514}.
\newblock \DOIprefix\doi{10.1007/978-3-319-24553-9_62}.
\bibitem[{Dai et~al.(2016)Dai, Li, He and Sun}]{Dai}
\bibinfo{author}{Dai, J.}, \bibinfo{author}{Li, Y.}, \bibinfo{author}{He, K.},
  \bibinfo{author}{Sun, J.}, \bibinfo{year}{2016}.
\newblock \bibinfo{title}{{R-FCN: Object detection via region-based fully
  convolutional networks}}, in: \bibinfo{booktitle}{Advances in Neural
  Information Processing Systems}, pp. \bibinfo{pages}{379--387}.
\newblock \URLprefix \url{https://github.com/daijifeng001/r-fcn.},
  \href{http://arxiv.org/abs/1605.06409}{\tt arXiv:1605.06409}.
\bibitem[{Deng and Cahill(1994)}]{Deng}
\bibinfo{author}{Deng, G.}, \bibinfo{author}{Cahill, L.W.},
  \bibinfo{year}{1994}.
\newblock \bibinfo{title}{{Adaptive Gaussian filter for noise reduction and
  edge detection}}, in: \bibinfo{booktitle}{IEEE Nuclear Science Symposium {\&}
  Medical Imaging Conference}, pp. \bibinfo{pages}{1615--1619}.
\newblock \URLprefix
  \url{https://ieeexplore.ieee.org/abstract/document/373563/},
  \DOIprefix\doi{10.1109/nssmic.1993.373563}.
\bibitem[{Fawcett(2006)}]{Letters}
\bibinfo{author}{Fawcett, T.}, \bibinfo{year}{2006}.
\newblock \bibinfo{title}{{An introduction to ROC analysis}}.
\newblock \bibinfo{journal}{Pattern Recognition Letters} \bibinfo{volume}{27},
  \bibinfo{pages}{861--874}.
\newblock \URLprefix
  \url{https://www.sciencedirect.com/science/article/pii/S016786550500303X},
  \DOIprefix\doi{10.1016/j.patrec.2005.10.010}.
\bibitem[{Ghesu et~al.(2019)Ghesu, Georgescu, Zheng, Grbic, Maier, Hornegger
  and Comaniciu}]{Ghesu}
\bibinfo{author}{Ghesu, F.C.}, \bibinfo{author}{Georgescu, B.},
  \bibinfo{author}{Zheng, Y.}, \bibinfo{author}{Grbic, S.},
  \bibinfo{author}{Maier, A.}, \bibinfo{author}{Hornegger, J.},
  \bibinfo{author}{Comaniciu, D.}, \bibinfo{year}{2019}.
\newblock \bibinfo{title}{{Multi-Scale Deep Reinforcement Learning for
  Real-Time 3D-Landmark Detection in CT Scans}}.
\newblock \bibinfo{journal}{IEEE Transactions on Pattern Analysis and Machine
  Intelligence} \bibinfo{volume}{41}, \bibinfo{pages}{176--189}.
\newblock \URLprefix
  \url{https://ieeexplore.ieee.org/abstract/document/8187667/},
  \DOIprefix\doi{10.1109/TPAMI.2017.2782687}.
\bibitem[{Ghesu et~al.(2016)Ghesu, Krubasik, Georgescu, Singh, Zheng, Hornegger
  and Comaniciu}]{Ghesu2016}
\bibinfo{author}{Ghesu, F.C.}, \bibinfo{author}{Krubasik, E.},
  \bibinfo{author}{Georgescu, B.}, \bibinfo{author}{Singh, V.},
  \bibinfo{author}{Zheng, Y.}, \bibinfo{author}{Hornegger, J.},
  \bibinfo{author}{Comaniciu, D.}, \bibinfo{year}{2016}.
\newblock \bibinfo{title}{{Marginal Space Deep Learning: Efficient Architecture
  for Volumetric Image Parsing}}.
\newblock \bibinfo{journal}{IEEE Transactions on Medical Imaging}
  \bibinfo{volume}{35}, \bibinfo{pages}{1217--1228}.
\newblock \URLprefix \url{http://ieeexplore.ieee.org.},
  \DOIprefix\doi{10.1109/TMI.2016.2538802}.
\bibitem[{He et~al.(2015)He, Zhang, Ren and Sun}]{He}
\bibinfo{author}{He, K.}, \bibinfo{author}{Zhang, X.}, \bibinfo{author}{Ren,
  S.}, \bibinfo{author}{Sun, J.}, \bibinfo{year}{2015}.
\newblock \bibinfo{title}{{Spatial Pyramid Pooling in Deep Convolutional
  Networks for Visual Recognition}}.
\newblock \bibinfo{journal}{IEEE Transactions on Pattern Analysis and Machine
  Intelligence} \bibinfo{volume}{37}, \bibinfo{pages}{1904--1916}.
\newblock \URLprefix
  \url{https://ieeexplore.ieee.org/abstract/document/7005506/},
  \DOIprefix\doi{10.1109/TPAMI.2015.2389824},
  \href{http://arxiv.org/abs/1406.4729}{\tt arXiv:1406.4729}.
\bibitem[{He et~al.(2016a)He, Zhang, Ren and Sun}]{He2016a}
\bibinfo{author}{He, K.}, \bibinfo{author}{Zhang, X.}, \bibinfo{author}{Ren,
  S.}, \bibinfo{author}{Sun, J.}, \bibinfo{year}{2016}a.
\newblock \bibinfo{title}{{Deep residual learning for image recognition}}, in:
  \bibinfo{booktitle}{Proceedings of the IEEE Computer Society Conference on
  Computer Vision and Pattern Recognition}.
\newblock \DOIprefix\doi{10.1109/CVPR.2016.90},
  \href{http://arxiv.org/abs/1512.03385}{\tt arXiv:1512.03385}.
\bibitem[{He et~al.(2016b)He, Zhang, Ren and Sun}]{He2016}
\bibinfo{author}{He, K.}, \bibinfo{author}{Zhang, X.}, \bibinfo{author}{Ren,
  S.}, \bibinfo{author}{Sun, J.}, \bibinfo{year}{2016}b.
\newblock \bibinfo{title}{{Identity mappings in deep residual networks}}, in:
  \bibinfo{booktitle}{Lecture Notes in Computer Science (including subseries
  Lecture Notes in Artificial Intelligence and Lecture Notes in
  Bioinformatics)}, \bibinfo{publisher}{Springer Verlag}. pp.
  \bibinfo{pages}{630--645}.
\newblock \DOIprefix\doi{10.1007/978-3-319-46493-0_38},
  \href{http://arxiv.org/abs/1603.05027}{\tt arXiv:1603.05027}.
\bibitem[{Hill et~al.(1990)Hill, Kislak and Martin}]{Hilla}
\bibinfo{author}{Hill, L.M.}, \bibinfo{author}{Kislak, S.},
  \bibinfo{author}{Martin, J.G.}, \bibinfo{year}{1990}.
\newblock \bibinfo{title}{{Transvaginal sonographic detection of the
  pseudogestational sac associated with ectopic pregnancy}}.
\newblock \bibinfo{journal}{Obstetrics and Gynecology} \bibinfo{volume}{75},
  \bibinfo{pages}{986--988}.
\newblock \URLprefix \url{https://europepmc.org/abstract/med/2188184}.
\bibitem[{Jeve et~al.(2011)Jeve, Rana, Bhide and Thangaratinam}]{Jeve2011a}
\bibinfo{author}{Jeve, Y.}, \bibinfo{author}{Rana, R.}, \bibinfo{author}{Bhide,
  A.}, \bibinfo{author}{Thangaratinam, S.}, \bibinfo{year}{2011}.
\newblock \bibinfo{title}{{Accuracy of first-trimester ultrasound in the
  diagnosis of early embryonic demise: A systematic review}}.
\newblock \DOIprefix\doi{10.1002/uog.10108}.
\bibitem[{Kebir and Mekaoui(2019)}]{Kebir2019}
\bibinfo{author}{Kebir, S.T.}, \bibinfo{author}{Mekaoui, S.},
  \bibinfo{year}{2019}.
\newblock \bibinfo{title}{{An Efficient Methodology of Brain Abnormalities
  Detection using CNN Deep Learning Network}}, in:
  \bibinfo{booktitle}{Proceedings of the 2018 International Conference on
  Applied Smart Systems, ICASS 2018}.
\newblock \URLprefix
  \url{https://ieeexplore.ieee.org/abstract/document/8652054/},
  \DOIprefix\doi{10.1109/ICASS.2018.8652054}.
\bibitem[{Krizhevsky et~al.(2012)Krizhevsky, Sutskever and
  Hinton}]{Krizhevsky2012}
\bibinfo{author}{Krizhevsky, A.}, \bibinfo{author}{Sutskever, I.},
  \bibinfo{author}{Hinton, G.E.}, \bibinfo{year}{2012}.
\newblock \bibinfo{title}{{ImageNet classification with deep convolutional
  neural networks}}, in: \bibinfo{booktitle}{Advances in Neural Information
  Processing Systems}.
\bibitem[{Kumar and Shriram(2015)}]{Kumar}
\bibinfo{author}{Kumar, A.M.}, \bibinfo{author}{Shriram, K.S.},
  \bibinfo{year}{2015}.
\newblock \bibinfo{title}{{Automated scoring of fetal abdomen ultrasound
  scan-planes for biometry}}, in: \bibinfo{booktitle}{Proceedings -
  International Symposium on Biomedical Imaging}, pp.
  \bibinfo{pages}{862--865}.
\newblock \URLprefix
  \url{https://ieeexplore.ieee.org/abstract/document/7164007/},
  \DOIprefix\doi{10.1109/ISBI.2015.7164007}.
\bibitem[{Lin et~al.(2013)Lin, Chen and Yan}]{Lin2013}
\bibinfo{author}{Lin, M.}, \bibinfo{author}{Chen, Q.}, \bibinfo{author}{Yan,
  S.}, \bibinfo{year}{2013}.
\newblock \bibinfo{title}{{Network In Network}} \URLprefix
  \url{http://arxiv.org/abs/1312.4400},
  \href{http://arxiv.org/abs/1312.4400}{\tt arXiv:1312.4400}.
\bibitem[{Lin et~al.(2017a)Lin, Doll{\'{a}}r, Girshick, He, Hariharan and
  Belongie}]{Lin2017a}
\bibinfo{author}{Lin, T.Y.}, \bibinfo{author}{Doll{\'{a}}r, P.},
  \bibinfo{author}{Girshick, R.}, \bibinfo{author}{He, K.},
  \bibinfo{author}{Hariharan, B.}, \bibinfo{author}{Belongie, S.},
  \bibinfo{year}{2017}a.
\newblock \bibinfo{title}{{Feature pyramid networks for object detection}}, in:
  \bibinfo{booktitle}{Proceedings - 30th IEEE Conference on Computer Vision and
  Pattern Recognition, CVPR 2017}, pp. \bibinfo{pages}{936--944}.
\newblock \URLprefix
  \url{http://openaccess.thecvf.com/content{\_}cvpr{\_}2017/html/Lin{\_}Feature{\_}Pyramid{\_}Networks{\_}CVPR{\_}2017{\_}paper.html},
  \DOIprefix\doi{10.1109/CVPR.2017.106},
  \href{http://arxiv.org/abs/1612.03144}{\tt arXiv:1612.03144}.
\bibitem[{Lin et~al.(2017b)Lin, Goyal, Girshick, He and Dollar}]{Lin2017}
\bibinfo{author}{Lin, T.Y.}, \bibinfo{author}{Goyal, P.},
  \bibinfo{author}{Girshick, R.}, \bibinfo{author}{He, K.},
  \bibinfo{author}{Dollar, P.}, \bibinfo{year}{2017}b.
\newblock \bibinfo{title}{{Focal Loss for Dense Object Detection}}, in:
  \bibinfo{booktitle}{Proceedings of the IEEE International Conference on
  Computer Vision}, \bibinfo{publisher}{Institute of Electrical and Electronics
  Engineers Inc.}. pp. \bibinfo{pages}{2999--3007}.
\newblock \DOIprefix\doi{10.1109/ICCV.2017.324},
  \href{http://arxiv.org/abs/1708.02002}{\tt arXiv:1708.02002}.
\bibitem[{Lin et~al.(2019)Lin, Li, Ni, Liao, Wen, Du, Chen, Wang and Lei}]{Lin}
\bibinfo{author}{Lin, Z.}, \bibinfo{author}{Li, S.}, \bibinfo{author}{Ni, D.},
  \bibinfo{author}{Liao, Y.}, \bibinfo{author}{Wen, H.}, \bibinfo{author}{Du,
  J.}, \bibinfo{author}{Chen, S.}, \bibinfo{author}{Wang, T.},
  \bibinfo{author}{Lei, B.}, \bibinfo{year}{2019}.
\newblock \bibinfo{title}{{Multi-task learning for quality assessment of fetal
  head ultrasound images}}.
\newblock \bibinfo{journal}{Medical Image Analysis} \bibinfo{volume}{58},
  \bibinfo{pages}{101548}.
\newblock \URLprefix
  \url{https://www.sciencedirect.com/science/article/pii/S1361841519300830},
  \DOIprefix\doi{10.1016/j.media.2019.101548}.
\bibitem[{Liu et~al.(2016)Liu, Anguelov, Erhan, Szegedy, Reed, Fu and
  Berg}]{Liu2016}
\bibinfo{author}{Liu, W.}, \bibinfo{author}{Anguelov, D.},
  \bibinfo{author}{Erhan, D.}, \bibinfo{author}{Szegedy, C.},
  \bibinfo{author}{Reed, S.}, \bibinfo{author}{Fu, C.Y.},
  \bibinfo{author}{Berg, A.C.}, \bibinfo{year}{2016}.
\newblock \bibinfo{title}{{SSD: Single shot multibox detector}}, in:
  \bibinfo{booktitle}{Lecture Notes in Computer Science (including subseries
  Lecture Notes in Artificial Intelligence and Lecture Notes in
  Bioinformatics)}, \bibinfo{publisher}{Springer Verlag}. pp.
  \bibinfo{pages}{21--37}.
\newblock \DOIprefix\doi{10.1007/978-3-319-46448-0_2},
  \href{http://arxiv.org/abs/1512.02325}{\tt arXiv:1512.02325}.
\bibitem[{Murphy et~al.(2018)Murphy, Xu, Kochanek and Arias}]{Murphy2018}
\bibinfo{author}{Murphy, S.L.}, \bibinfo{author}{Xu, J.},
  \bibinfo{author}{Kochanek, K.D.}, \bibinfo{author}{Arias, E.},
  \bibinfo{year}{2018}.
\newblock \bibinfo{title}{{Mortality in the United States, 2017 Key findings
  Data from the National Vital Statistics System}} ,
  \bibinfo{pages}{1--8}\URLprefix
  \url{https://www.cdc.gov/nchs/products/databriefs/db328.htm}.
\bibitem[{Namburete et~al.(2018)Namburete, Xie, Yaqub, Zisserman and
  Noble}]{Namburete2018}
\bibinfo{author}{Namburete, A.I.}, \bibinfo{author}{Xie, W.},
  \bibinfo{author}{Yaqub, M.}, \bibinfo{author}{Zisserman, A.},
  \bibinfo{author}{Noble, J.A.}, \bibinfo{year}{2018}.
\newblock \bibinfo{title}{{Fully-automated alignment of 3D fetal brain
  ultrasound to a canonical reference space using multi-task learning}}.
\newblock \bibinfo{journal}{Medical Image Analysis} \bibinfo{volume}{46},
  \bibinfo{pages}{1--14}.
\newblock \URLprefix \url{https://doi.org/10.1016/j.media.2018.02.006},
  \DOIprefix\doi{10.1016/j.media.2018.02.006}.
\bibitem[{Redmon et~al.(2016)Redmon, Divvala, Girshick and
  Farhadi}]{Redmon2016}
\bibinfo{author}{Redmon, J.}, \bibinfo{author}{Divvala, S.},
  \bibinfo{author}{Girshick, R.}, \bibinfo{author}{Farhadi, A.},
  \bibinfo{year}{2016}.
\newblock \bibinfo{title}{{You only look once: Unified, real-time object
  detection}}, in: \bibinfo{booktitle}{Proceedings of the IEEE Computer Society
  Conference on Computer Vision and Pattern Recognition}, pp.
  \bibinfo{pages}{779--788}.
\newblock \URLprefix \url{http://pjreddie.com/yolo/},
  \DOIprefix\doi{10.1109/CVPR.2016.91},
  \href{http://arxiv.org/abs/1506.02640}{\tt arXiv:1506.02640}.
\bibitem[{Redmon and Farhadi()}]{Redmon}
\bibinfo{author}{Redmon, J.}, \bibinfo{author}{Farhadi, A.}, .
\newblock \bibinfo{title}{{YOLO9000: Better, Faster, Stronger}}.
\newblock \bibinfo{type}{Technical Report}.
\newblock \URLprefix \url{http://pjreddie.com/yolo9000/}.
\bibitem[{Ren et~al.(2017)Ren, He, Girshick and Sun}]{Ren}
\bibinfo{author}{Ren, S.}, \bibinfo{author}{He, K.}, \bibinfo{author}{Girshick,
  R.}, \bibinfo{author}{Sun, J.}, \bibinfo{year}{2017}.
\newblock \bibinfo{title}{{Faster R-CNN: Towards Real-Time Object Detection
  with Region Proposal Networks}}.
\newblock \bibinfo{journal}{IEEE Transactions on Pattern Analysis and Machine
  Intelligence} \bibinfo{volume}{39}, \bibinfo{pages}{1137--1149}.
\newblock \URLprefix \url{https://github.com/},
  \DOIprefix\doi{10.1109/TPAMI.2016.2577031},
  \href{http://arxiv.org/abs/1506.01497}{\tt arXiv:1506.01497}.
\bibitem[{Rueda et~al.(2014)Rueda, Fathima, Knight, Yaqub, Papageorghiou,
  Rahmatullah, Foi, Maggioni, Pepe, Tohka, Stebbing, McManigle, Ciurte,
  Bresson, Cuadra, Sun, Ponomarev, Gelfand, Kazanov, Wang, Chen, Peng, Hung and
  Noble}]{Rueda}
\bibinfo{author}{Rueda, S.}, \bibinfo{author}{Fathima, S.},
  \bibinfo{author}{Knight, C.L.}, \bibinfo{author}{Yaqub, M.},
  \bibinfo{author}{Papageorghiou, A.T.}, \bibinfo{author}{Rahmatullah, B.},
  \bibinfo{author}{Foi, A.}, \bibinfo{author}{Maggioni, M.},
  \bibinfo{author}{Pepe, A.}, \bibinfo{author}{Tohka, J.},
  \bibinfo{author}{Stebbing, R.V.}, \bibinfo{author}{McManigle, J.E.},
  \bibinfo{author}{Ciurte, A.}, \bibinfo{author}{Bresson, X.},
  \bibinfo{author}{Cuadra, M.B.}, \bibinfo{author}{Sun, C.},
  \bibinfo{author}{Ponomarev, G.V.}, \bibinfo{author}{Gelfand, M.S.},
  \bibinfo{author}{Kazanov, M.D.}, \bibinfo{author}{Wang, C.W.},
  \bibinfo{author}{Chen, H.C.}, \bibinfo{author}{Peng, C.W.},
  \bibinfo{author}{Hung, C.M.}, \bibinfo{author}{Noble, J.A.},
  \bibinfo{year}{2014}.
\newblock \bibinfo{title}{{Evaluation and comparison of current fetal
  ultrasound image segmentation methods for biometric measurements: A grand
  challenge}}.
\newblock \bibinfo{journal}{IEEE Transactions on Medical Imaging}
  \bibinfo{volume}{33}, \bibinfo{pages}{797--813}.
\newblock \URLprefix
  \url{https://ieeexplore.ieee.org/abstract/document/6575204/},
  \DOIprefix\doi{10.1109/TMI.2013.2276943}.
\bibitem[{Simonyan and Zisserman(2014)}]{Simonyan2014}
\bibinfo{author}{Simonyan, K.}, \bibinfo{author}{Zisserman, A.},
  \bibinfo{year}{2014}.
\newblock \bibinfo{title}{{Very Deep Convolutional Networks for Large-Scale
  Image Recognition}} , \bibinfo{pages}{1--14}\URLprefix
  \url{http://arxiv.org/abs/1409.1556},
  \href{http://arxiv.org/abs/1409.1556}{\tt arXiv:1409.1556}.
\bibitem[{Sujit et~al.(2019)Sujit, Gabr, Coronado, Robinson, Datta and
  Narayana}]{Sujit}
\bibinfo{author}{Sujit, S.J.}, \bibinfo{author}{Gabr, R.E.},
  \bibinfo{author}{Coronado, I.}, \bibinfo{author}{Robinson, M.},
  \bibinfo{author}{Datta, S.}, \bibinfo{author}{Narayana, P.A.},
  \bibinfo{year}{2019}.
\newblock \bibinfo{title}{{Automated Image Quality Evaluation of Structural
  Brain Magnetic Resonance Images using Deep Convolutional Neural Networks}},
  in: \bibinfo{booktitle}{2018 9th Cairo International Biomedical Engineering
  Conference, CIBEC 2018 - Proceedings}, pp. \bibinfo{pages}{33--36}.
\newblock \URLprefix
  \url{https://ieeexplore.ieee.org/abstract/document/8641830/},
  \DOIprefix\doi{10.1109/CIBEC.2018.8641830}.
\bibitem[{Thilaganathan(2011)}]{Thilaganathan2011}
\bibinfo{author}{Thilaganathan, B.}, \bibinfo{year}{2011}.
\newblock \bibinfo{title}{{Opinion: The evidence base for miscarriage
  diagnosis: Better late than never}}.
\newblock \DOIprefix\doi{10.1002/uog.10110}.
\bibitem[{Wu et~al.(2017)Wu, Cheng, Li, Lei, Wang and Ni}]{Wu2017}
\bibinfo{author}{Wu, L.}, \bibinfo{author}{Cheng, J.Z.}, \bibinfo{author}{Li,
  S.}, \bibinfo{author}{Lei, B.}, \bibinfo{author}{Wang, T.},
  \bibinfo{author}{Ni, D.}, \bibinfo{year}{2017}.
\newblock \bibinfo{title}{{FUIQA: Fetal ultrasound image quality assessment
  with deep convolutional networks}}.
\newblock \bibinfo{journal}{IEEE Transactions on Cybernetics}
  \bibinfo{volume}{47}, \bibinfo{pages}{1336--1349}.
\newblock \URLprefix
  \url{http://www.ieee.org/publications{\_}standards/publications/rights/index.html},
  \DOIprefix\doi{10.1109/TCYB.2017.2671898}.
\bibitem[{Xu et~al.(2018)Xu, Huo, Park, Landman, Milkowski, Grbic and
  Zhou}]{Xu2018}
\bibinfo{author}{Xu, Z.}, \bibinfo{author}{Huo, Y.}, \bibinfo{author}{Park,
  J.H.}, \bibinfo{author}{Landman, B.}, \bibinfo{author}{Milkowski, A.},
  \bibinfo{author}{Grbic, S.}, \bibinfo{author}{Zhou, S.},
  \bibinfo{year}{2018}.
\newblock \bibinfo{title}{{Less is more: Simultaneous view classification and
  landmark detection for abdominal ultrasound images}}, in:
  \bibinfo{booktitle}{Lecture Notes in Computer Science (including subseries
  Lecture Notes in Artificial Intelligence and Lecture Notes in
  Bioinformatics)}, \bibinfo{publisher}{Springer Verlag}. pp.
  \bibinfo{pages}{711--719}.
\newblock \DOIprefix\doi{10.1007/978-3-030-00934-2_79},
  \href{http://arxiv.org/abs/1805.10376}{\tt arXiv:1805.10376}.
\bibitem[{Zhang et~al.(2017a)Zhang, Liu and Shen}]{Zhanga}
\bibinfo{author}{Zhang, J.}, \bibinfo{author}{Liu, M.}, \bibinfo{author}{Shen,
  D.}, \bibinfo{year}{2017}a.
\newblock \bibinfo{title}{{Detecting Anatomical Landmarks from Limited Medical
  Imaging Data Using Two-Stage Task-Oriented Deep Neural Networks}}.
\newblock \bibinfo{journal}{IEEE Transactions on Image Processing}
  \bibinfo{volume}{26}, \bibinfo{pages}{4753--4764}.
\newblock \URLprefix
  \url{https://ieeexplore.ieee.org/abstract/document/7961205/},
  \DOIprefix\doi{10.1109/TIP.2017.2721106}.
\bibitem[{Zhang et~al.(2017b)Zhang, Dudley, Lambrou, Allinson and Ye}]{Zhang}
\bibinfo{author}{Zhang, L.}, \bibinfo{author}{Dudley, N.J.},
  \bibinfo{author}{Lambrou, T.}, \bibinfo{author}{Allinson, N.},
  \bibinfo{author}{Ye, X.}, \bibinfo{year}{2017}b.
\newblock \bibinfo{title}{{Automatic image quality assessment and measurement
  of fetal head in two-dimensional ultrasound image}}.
\newblock \bibinfo{journal}{Journal of Medical Imaging} \bibinfo{volume}{4},
  \bibinfo{pages}{024001}.
\newblock \URLprefix
  \url{https://www.spiedigitallibrary.org/journals/Journal-of-Medical-Imaging/volume-4/issue-2/024001/Automatic-image-quality-assessment-and-measurement-of-fetal-head-in/10.1117/1.JMI.4.2.024001.short},
  \DOIprefix\doi{10.1117/1.jmi.4.2.024001}.
\bibitem[{Zhao et~al.(2019)Zhao, Liu and Fingscheidt}]{Zhao}
\bibinfo{author}{Zhao, Z.}, \bibinfo{author}{Liu, H.},
  \bibinfo{author}{Fingscheidt, T.}, \bibinfo{year}{2019}.
\newblock \bibinfo{title}{{Convolutional neural networks to enhance coded
  speech}}.
\newblock \bibinfo{journal}{IEEE/ACM Transactions on Audio Speech and Language
  Processing} \bibinfo{volume}{27}, \bibinfo{pages}{663--678}.
\newblock \URLprefix
  \url{https://ieeexplore.ieee.org/abstract/document/8579579/},
  \DOIprefix\doi{10.1109/TASLP.2018.2887337},
  \href{http://arxiv.org/abs/1806.09411}{\tt arXiv:1806.09411}.

\end{thebibliography}

\end{document}